\documentclass{article}

\usepackage{microtype}
\usepackage{graphicx}
\usepackage{subfigure}
\usepackage{booktabs}
\usepackage{amsmath}
\usepackage{amssymb}

\usepackage{hyperref}
\usepackage[capitalise]{cleveref}

\makeatletter
\newcommand\footnoteref[1]{\protected@xdef\@thefnmark{\ref{#1}}\@footnotemark}
\makeatother

\usepackage[accepted]{icml2018}

\DeclareMathOperator{\KL}{KL}
\icmltitlerunning{A Hierarchical Latent Vector Model for Learning Long-Term Structure in Music}

\begin{document}

\twocolumn[
\icmltitle{A Hierarchical Latent Vector Model\\ for Learning Long-Term Structure in Music}

\icmlsetsymbol{equal}{*}

\begin{icmlauthorlist}
\icmlauthor{Adam Roberts}{goo}
\icmlauthor{Jesse Engel}{goo}
\icmlauthor{Colin Raffel}{goo}
\icmlauthor{Curtis Hawthorne}{goo}
\icmlauthor{Douglas Eck}{goo}
\end{icmlauthorlist}

\icmlaffiliation{goo}{Google Brain, Mountain View, CA, USA}

\icmlcorrespondingauthor{Adam Roberts}{adarob@google.com}

\icmlkeywords{Generative Modeling,Latent Variable Models,Hierarchical Recurrent Neural Networks,Music Generation}

\vskip 0.3in
]

\printAffiliationsAndNotice{}  

\setcounter{footnote}{1}

\begin{abstract}

The Variational Autoencoder (VAE) has proven to be an effective model for producing semantically meaningful latent representations for natural data.
However, it has thus far seen limited application to sequential data, and, as we demonstrate, existing recurrent VAE models have difficulty modeling sequences with long-term structure.
To address this issue, we propose the use of a \textit{hierarchical} decoder, which first outputs embeddings for subsequences of the input and then uses these embeddings to generate each subsequence independently.
This structure encourages the model to utilize its latent code, thereby avoiding the ``posterior collapse'' problem, which remains an issue for recurrent VAEs.
We apply this architecture to modeling sequences of musical notes and find that it exhibits dramatically better sampling, interpolation, and reconstruction performance than a ``flat'' baseline model.
An implementation of our ``MusicVAE'' is available online.\footnote{\label{fn:code}\url{https://goo.gl/magenta/musicvae-code}}

\end{abstract}

\section{Introduction}

\begin{figure}[htb]
\vskip 0.0in
\begin{center}
\centerline{\includegraphics[width=\columnwidth]{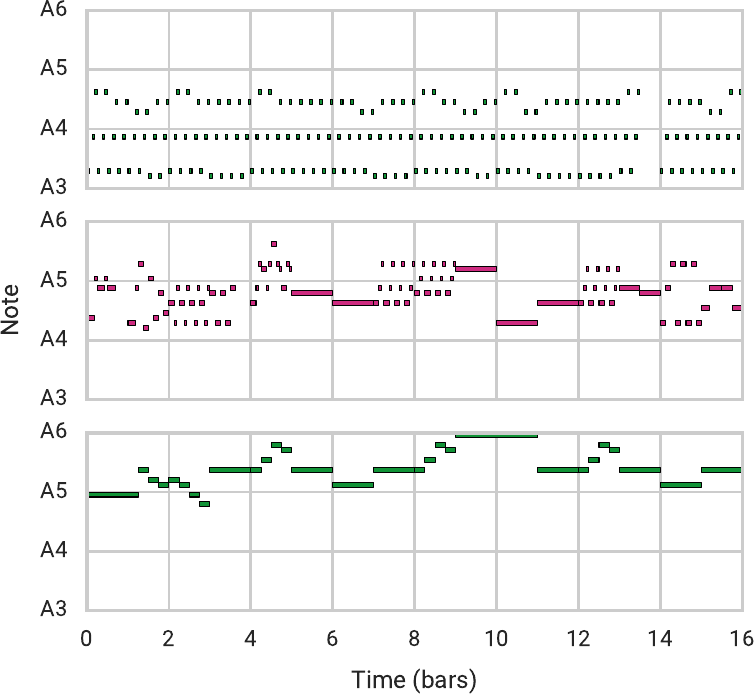}}
\caption{
Demonstration of latent-space averaging using MusicVAE.
The latent codes for the top and bottom sequences are averaged and decoded by our model to produce the middle sequence.
The latent-space mean involves a similar repeating pattern to the top sequence, but in a higher register and with intermittent pauses like the bottom sequence.
Audio for this example is available in the online supplement.\footnoteref{fn:mainonlinesupp} See \cref{fig:interpolations_data,fig:interpolations_hierarchical} in \cref{sec:additional_figures} for a baseline comparison.
}
\label{fig:interp_mean}
\end{center}
\vskip -0.2in
\end{figure}

\stepcounter{footnote}\footnotetext{\label{fn:mainonlinesupp}\url{https://goo.gl/magenta/musicvae-examples}}

Generative modeling describes the framework of estimating the underlying probability distribution $p(x)$ used to generate data $x$.
This can facilitate a wide range of applications, from sampling novel datapoints to unsupervised representation learning to estimating the probability of an existing datapoint under the learned distribution.
Much recent progress in generative modeling has been expedited by the use of deep neural networks, producing ``deep generative models,'' which leverage the expressive power of deep networks to model complex and high-dimensional distributions.
Practical achievements include generating realistic images with millions of pixels \cite{karras2017progressive}, generating synthetic audio with hundreds of thousands of timesteps \cite{van2016wavenet}, and achieving state-of-the-art performance on semi-supervised learning tasks \cite{wei2018improving}.

A wide variety of methods have been used in deep generative modeling, including implicit models such as Generative Adversarial Networks (GANs) \cite{goodfellow2014generative} and explicit deep autoregressive models such as PixelCNN \cite{van2016conditional} and WaveNet \cite{van2016wavenet}.
In this work, we focus on deep latent variable models such as the Variational Autoencoder (VAE) \cite{kingma2014auto,rezende2014stochastic}.
The advantage of these models is that they explicitly model both $p(z | x)$ and $p(z)$, where $z$ is a latent vector that can either be inferred from existing data or sampled from a distribution over the latent space.
Ideally, the latent vector captures the pertinent characteristics of a given datapoint and disentangles factors of variation in a dataset.
These autoencoders also model the likelihood $p(x | z)$, which provides an efficient way of mapping the latent vector back to the data space.

Our interest in deep latent variable models primarily comes from their increasing use in creative applications of machine learning \cite{carter2017aia,ha2017neural,engel2017latent}.
This arises from surprising and convenient characteristics of the latent spaces typically learned by these models.
For example, averaging the latent codes for all datapoints that possess a given attribute produces a so-called attribute vector, which can be used to make targeted changes to data examples.
Encoding a datapoint with some attribute (say, a photograph of a person with brown hair) to obtain its latent code, subtracting the corresponding attribute vector (the ``brown hair'' vector), adding another attribute vector (``blond hair''), and decoding the resulting latent vector can produce a realistic manifestation of the initial point with the attributes swapped (the same person with blond hair) \cite{larsen2016autoencoding,mikolov2013distributed}.
As another example, interpolating between latent vectors and decoding points on the trajectory can produce realistic intermediate datapoints that morph between the characteristics of the ends in a smooth and semantically meaningful way.

Most work on deep latent variable models has focused on continuous-valued data with a fixed dimensionality, e.g., images.
Modeling sequential data is less common, particularly sequences of discrete tokens such as musical scores, which typically require the use of an autoregressive decoder.
This is partially because autoregression is often sufficiently powerful that the autoencoder ignores the latent code \cite{bowman2016generating}.
While they have shown some success on short sequences (e.g., sentences), deep latent variable models have yet to be successfully applied to very long sequences.

To address this gap, we introduce a novel sequential autoencoder with a hierarchical recurrent decoder, which helps overcome the aforementioned issue of modeling long-term structure with recurrent VAEs.
Our model encodes an entire sequence to a single latent vector, which enables many of the creative applications enjoyed by VAEs of images.
We show experimentally that our model is capable of effectively autoencoding substantially longer sequences than a baseline model with a ``flat'' decoder RNN.

In this paper, we focus on the application of modeling sequences of musical notes.
Western popular music exhibits strong long-term structure, such as the repetition and variation between measures and sections of a piece of music.
This structure is also hierarchical--songs are divided into sections, which are broken up into measures, and then into beats, and so on.
Further, music is fundamentally a multi-stream signal, in the sense that it often involves multiple players with strong inter-player dependencies.
These unique properties, in addition to the potential creative applications, make music an ideal testbed for our sequential autoencoder.

After covering a background of work our approach builds on, we describe our model and its novel architectural enhancements.
We then provide an overview of related work on applying latent variable models to sequences.
Finally, we demonstrate the ability of our method to model musical data through quantitative and qualitative evaluations.

\section{Background}
\label{sec:background}

Fundamentally, our model is an autoencoder, i.e., its goal is to accurately reconstruct its inputs.
However, we additionally desire the ability to draw novel samples and perform latent-space interpolations and attribute vector arithmetic.
For these properties, we adopt the framework of the Variational Autoencoder.
Successfully using VAEs for sequences benefits from some additional extensions to the VAE objective.
In the following subsections, we cover the prior work that forms the backbone for our approach.

\subsection{Variational Autoencoders}

A common constraint applied to autoencoders is that they compress the relevant information about the input into a lower-dimensional latent code.
Ideally, this forces the model to produce a compressed representation that captures important factors of variation in the dataset.
In pursuit of this goal, the Variational Autoencoder \cite{kingma2014auto,rezende2014stochastic} introduces the constraint that the latent code $z$ is a random variable distributed according to a prior $p(z)$.
The data generation model is then $z \sim p(z), x \sim p(x | z)$.
The VAE consists of an encoder $q_\lambda(z | x)$, which approximates the posterior $p(z | x)$, and a decoder $p_\theta(x | z)$, which parameterizes the likelihood $p(x | z)$.
In practice, the approximate posterior and likelihood distributions (``encoder'' and ``decoder'') are parameterized by neural networks with parameters $\lambda$ and $\theta$ respectively.
Following the framework of Variational Inference, we do posterior inference by minimizing the KL divergence between our approximate posterior, the encoder, and the true posterior $p(z | x)$ by maximizing the evidence lower bound (ELBO)
\begin{equation}
        \mathbb{E}[ \log p_\theta(x | z) ] - \KL(q_\lambda(z | x) || p(z)) \le \log p(x)
\end{equation}
where the expectation is taken with respect to $z \sim q_\lambda(z | x)$ and $\KL(\cdot || \cdot)$ is the KL-divergence.
Naively computing the gradient through the ELBO is infeasible due to the sampling operation used to obtain $z$.
In the common case where $p(z)$ is a diagonal-covariance Gaussian, this can be circumvented by replacing $z \sim \mathcal{N}(\mu, \sigma I)$ with
\begin{equation}
        \epsilon \sim \mathcal{N}(0, I), z = \mu + \sigma\odot\epsilon
\label{eq:z_sample}
\end{equation}


\subsubsection{$\beta$-VAE and Free Bits}

One way of interpreting the ELBO used in the VAE is by considering its two terms, $\mathbb{E}[ \log p_\theta(x | z) ]$ and $\KL(q_\lambda(z | x) || p(z))$, separately.
The first can be thought of as requiring that $p(x | z)$ is high for samples of $z$ from $q_\lambda(z | x)$--ensuring accurate reconstructions.
The second encourages $q_\lambda(z | x)$ to be close to the prior--ensuring we can generate realistic data by sampling latent codes from $p(z)$.  
The presence of these terms suggests a trade-off between the quality of samples and reconstructions--or equivalently, between the rate (amount of information encoded in $q_\lambda(z | x)$) and distortion (data likelihood) \cite{alemi2017information}.

As is, the ELBO has no way of directly controlling this trade-off.
A common modification to the ELBO introduces the KL weight hyperparameter $\beta$ \cite{bowman2016generating,higgins2016beta} producing
\begin{equation}
        \mathbb{E}[ \log p_\theta(x | z) ] - \beta \KL(q_\lambda(z | x) || p(z))
\end{equation}
Setting $\beta < 1$ encourages the model to prioritize reconstruction quality over learning a compact representation.

Another approach for adjusting this trade-off is to only enforce the KL regularization term once it exceeds a threshold \cite{kingma2016improved}:
\begin{equation}
        \mathbb{E}[ \log p_\theta(x | z) ] - \max(\KL(q_\lambda(z | x) || p(z)) - \tau, 0)
\end{equation}
This stems from the interpretation that $\KL(q_\lambda(z | x) || p(z))$ measures the amount of information required to code samples from $p(z)$ using $q_\lambda(z | x)$.
Utilizing this threshold therefore amounts to giving the model a certain budget of ``free bits'' to use when learning the approximate posterior.
Note that these modified objectives no longer optimize a lower bound on the likelihood, but as is custom we still refer to the resulting models as ``Variational Autoencoders.''

\subsubsection{Latent Space Manipulation}

The broad goal of an autoencoder is to learn a compact representation of the data.
For creative applications, we have additional uses for the latent space learned by the model \cite{carter2017aia,roberts2018milc}.
First, given a point in latent space that maps to a realistic datapoint, nearby latent space points should map to datapoints that are semantically similar.
By extrapolation, this implies that all points along a continuous curve connecting two points in latent space should be decodable to a series of datapoints that produce a smooth semantic interpolation in data space.
Further, this requirement effectively mandates that the latent space is ``smooth'' and does not contain any ``holes,'' i.e., isolated regions that do not map to realistic datapoints.
Second, we desire that the latent space disentangles meaningful semantic groups in the dataset.

Ideally, these requirements should be satisfied by a VAE if the likelihood and KL divergence terms are both sufficiently small on held-out test data.
A more practical test of these properties is to interpolate between points in the latent space and test whether the corresponding points in the data space are interpolated in a semantically meaningful way.
Concretely, if $z_1$ and $z_2$ are the latent vectors corresponding to datapoints $x_1$ and $x_2$, then we can perform linear interpolation in latent space by computing
\begin{equation}
\label{eq:lerp}
c_\alpha = \alpha z_1 + (1 - \alpha)z_2
\end{equation}
for $\alpha \in [0, 1]$.
Our goal is satisfied if $p_\theta(x | c_\alpha)$ is a realistic datapoint for all $\alpha$, $p_\theta(x | c_\alpha)$ transitions in a semantically meaningful way from $p_\theta(x | c_0)$ to $p_\theta(x | c_1)$ as we vary $\alpha$ from 0 to 1, and that $p_\theta(x | c_\alpha)$ is perceptually similar to $p_\theta(x | c_{\alpha + \delta})$ for small $\delta$.
Note that because the prior over the latent space of a VAE is a spherical Gaussian, samples from high-dimensional priors are practically indistinguishable from samples from the uniform distribution on the unit hypersphere \cite{huszar2017gaussian}.
In practice we therefore use spherical interpolation \cite{white2016sampling} instead of \cref{eq:lerp}.

An additional test for whether our autoencoder will be useful in creative applications measures whether it produces reliable ``attribute vectors.''
Attribute vectors are computed as the average latent vector for a collection of datapoints that share some particular attribute.
Typically, attribute vectors are computed for pairs of attributes, e.g., images of people with and without glasses.
The model's ability to discover attributes is then tested by encoding a datapoint with attribute A, subtracting the ``attribute A vector'' from its latent code, adding the ``attribute B vector'', and testing whether the decoded result appears like the original datapoint with attribute B instead of A.
In our experiments, we use the above latent space manipulation techniques to demonstrate the power of our proposed model.

\subsection{Recurrent VAEs}

While a wide variety of neural network structures have been considered for the encoder and decoder network in a VAE, in the present work we are most interested in models with a recurrent encoder and decoder \cite{bowman2016generating}. 
Concretely, the encoder, $q_\lambda(z | x)$, is a recurrent neural network (RNN) that processes the input sequence $x = \{x_1, x_2, \ldots, x_T\}$ and produces a sequence of hidden states $h_1, h_2, \ldots, h_T$.
The parameters of the distribution over the latent code $z$ are then set as a function of $h_T$.
The decoder, $p_\theta(x|z)$, uses the sampled latent vector $z$ to set the initial state of a decoder RNN, which autoregressively produces the output sequence $y = \{y_1, y_2, \ldots, y_T\}$.
The model is trained both to reconstruct the input sequence (i.e., $y_i = x_i, i \in \{1, \ldots, T\}$) and to learn an approximate posterior $q_\lambda(z | x)$ close to the prior $p(z)$, as in a standard VAE.

There are two main drawbacks of this approach.
First, RNNs are themselves typically used on their own as powerful autoregressive models of sequences.
As a result, the decoder in a recurrent VAE is itself sufficiently powerful to produce an effective model of the data and may disregard the latent code.
With the latent code ignored, the KL divergence term of the ELBO can be trivially set to zero, despite the fact that the model is effectively no longer acting as an autoencoder.
Second, the model must compress the entire sequence to a single latent vector.
While this approach has been shown to work for short sequences \cite{bowman2016generating,sutskever2014sequence}, it begins to fail as the sequence length increases \cite{bahdanau2014neural}.
In the following section, we present a latent variable autoencoder that overcomes these issues by using a hierarchical RNN for the decoder.

\begin{figure}[t]
\vskip 0.0in
\begin{center}
\centerline{\includegraphics[width=\columnwidth]{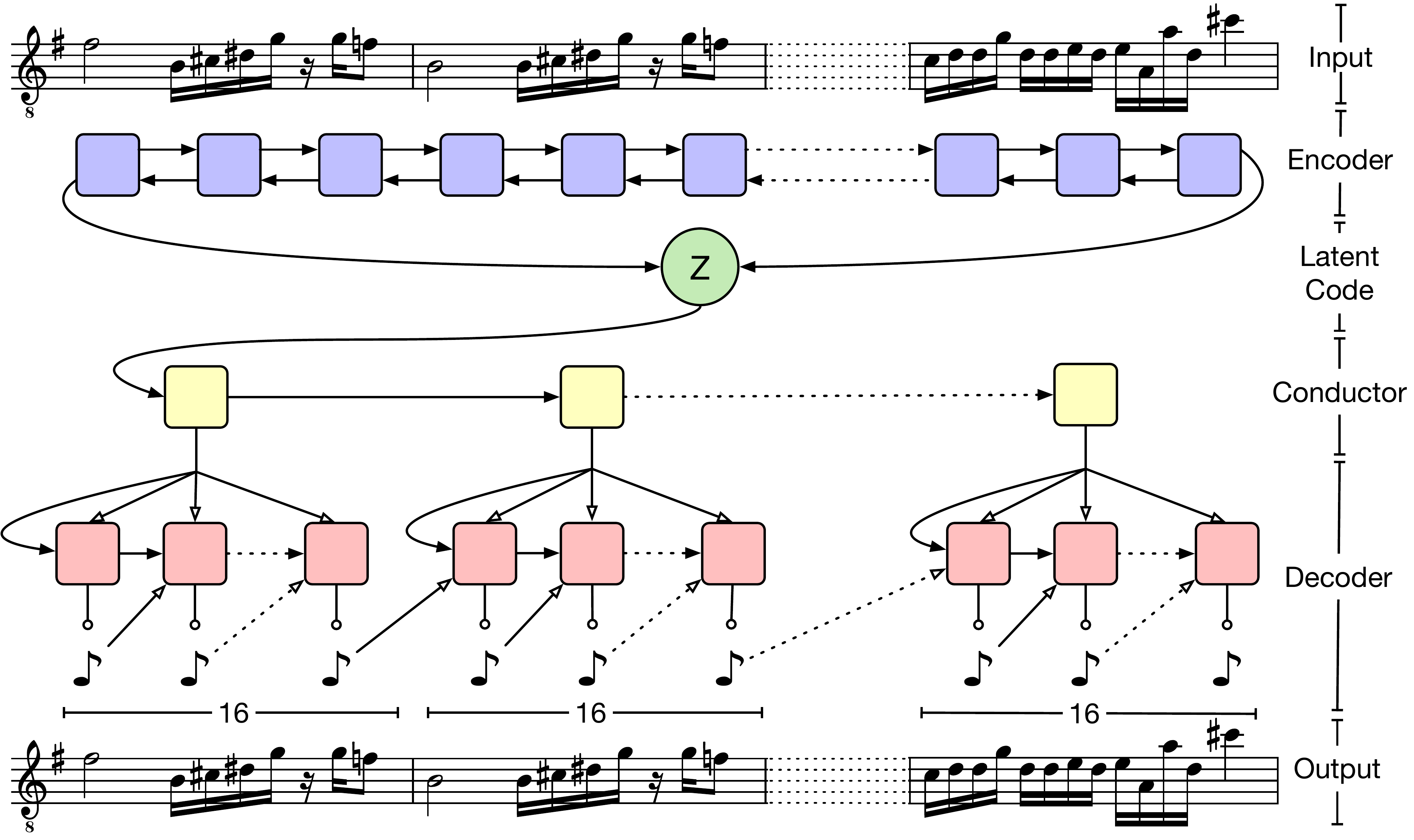}}
\caption{
Schematic of our hierarchical recurrent Variational Autoencoder model, MusicVAE.
}
\label{fig:model}
\end{center}
\vskip -0.4in
\end{figure}

\section{Model}
\label{sec:model}

At a high level, our model follows the basic structure used in previously-proposed VAEs for sequential data \cite{bowman2016generating}.
However, we introduce a novel hierarchical decoder, which we demonstrate produces substantially better performance on long sequences in \cref{sec:experiments}.
A schematic of our model, which we dub ``MusicVAE,'' is shown in \cref{fig:model}.

\subsection{Bidirectional Encoder}
\label{sec:encoder}

For the encoder $q_\lambda(z | x)$, we use a two-layer bidirectional LSTM network \cite{hochreiter1997long, schuster1997bidirectional}.
We process an input sequence $\mathbf{x} = \{x_1, x_2, \ldots, x_T\}$ to obtain the final state vectors $\overrightarrow{h}_T, \overleftarrow{h}_T$ from the second bidirectional LSTM layer.
These are then concatenated to produce $h_T$ and fed into two fully-connected layers to produce the latent distribution parameters $\mu$ and $\sigma$:
\begin{align}
    \label{eq:mu}
    \mu &= W_{h\mu} h_T + b_\mu\\
    \sigma &= \log\left(\exp(W_{h\sigma} h_T + b_\sigma) + 1\right)
\end{align}
where $W_{h\mu}, W_{h\sigma}$ and $b_\mu, b_\sigma$ are weight matrices and bias vectors, respectively.
In our experiments, we use an LSTM state size of 2048 for all layers and 512 latent dimensions.
As is standard in VAEs, $\mu$ and $\sigma$ then parametrize the latent distribution as in \cref{eq:z_sample}.
The use of a bidirectional recurrent encoder ideally gives the parametrization of the latent distribution longer-term context about the input sequence.

\subsection{Hierarchical Decoder}
\label{sec:decoder}

In prior work, the decoder in a recurrent VAE is typically a simple stacked RNN.
The decoder RNN uses the latent vector $z$ to set its initial state, and proceeds to generate the output sequence autoregressively.
In preliminary experiments (discussed in \cref{sec:experiments}), we found that using a simple RNN as the decoder resulted in poor sampling and reconstruction for long sequences.
We believe this is caused by the vanishing influence of the latent state as the output sequence is generated.

To mitigate this issue, we propose a novel hierarchical RNN for the decoder.
Assume that the input sequence (and target output sequence) $\mathbf{x}$ can be segmented into $U$ nonoverlapping subsequences $y_u$ with endpoints $i_u$ so that
\begin{align}
    y_u &= \{x_{i_u}, x_{i_u + 1}, x_{i_u + 2}, \ldots, x_{i_{u + 1} - 1}\} \\
    \rightarrow \mathbf{x} &= \{y_1, y_2, \ldots, y_U\}
\end{align}
where we define the special case of $i_{U + 1} = T$.
Then, the latent vector $z$ is passed through a fully-connected layer\footnote{Throughout, whenever we refer to a ``fully-connected layer,'' we mean a simple affine transformation as in \cref{eq:mu}.} followed by a $\tanh$ activation to get the initial state of a ``conductor'' RNN.
The conductor RNN produces $U$ embedding vectors $\mathbf{c} = \{c_1, c_2, \ldots, c_U\}$, one for each subsequence.
In our experiments, we use a two-layer unidirectional LSTM for the conductor with a hidden state size of 1024 and 512 output dimensions.

Once the conductor has produced the sequence of embedding vectors $\mathbf{c}$, each one is individually passed through a shared fully-connected layer followed by a $\tanh$ activation to produce initial states for a final bottom-layer decoder RNN.
The decoder RNN then autoregressively produces a sequence of distributions over output tokens for each subsequence $y_u$ via a softmax output layer.
At each step of the bottom-level decoder, the current conductor embedding $c_u$ is concatenated with the previous output token to be used as the input.
In our experiments, we used a 2-layer LSTM with 1024 units per layer for the decoder RNN.

In principle, our use of an autoregressive RNN decoder still allows for the ``posterior collapse'' problem where the model effectively learns to ignore the latent state.
Simliar to \cite{chen2016variational}, we find that it is important to limit the scope of the decoder to force it to use the latent code to model long-term structure. 
For a CNN decoder, this is as simple as reducing the receptive field, but no direct analogy exists for RNNs, which in principle have an unlimited temporal receptive field.
To get around this, we reduce the effective scope of the bottom-level RNN in the decoder by only allowing it to propagate state within an output subsequence. 
As described above, we initialize each subsequence RNN state with the corresponding embedding passed down by the conductor.
This implies that the only way for the decoder to get longer-term context is by using the embeddings produced by the conductor, which in turn depend solely on the latent code.
We experimented with an autoregressive version of the conductor where the decoder state was passed back to the conductor at the end of each subsequence, but found it exhibited worse performance.
We believe that these combined constraints effectively force the model to utilize the conductor embeddings, and by extension the latent vector, in order to correctly autoencode the sequence.

\subsection{Multi-Stream Modeling}

Many common sources of sequential data, such as text, consist solely of a single ``stream,'' i.e., there is only one sequence source which is producing tokens.
However, music is often a fundamentally multi-stream signal--a given musical sequence may consist of multiple players producing notes in tandem.
Modeling music therefore may also involve modeling the complex inter-stream dependencies.

We explore this possibility by introducing a ``trio'' model, which is identical to our basic MusicVAE except that it produces 3 separate distributions over output tokens--one for each of three instruments (drum, bass, and melody).
In our hierarchical decoder model, we consider these separate streams as an orthogonal ``dimension'' of hierarchy, and use a separate decoder RNN for each instrument.
The embeddings from the conductor RNN initialize the states of each instrument RNN through separate fully-connected layers followed by $\tanh$ activations.
For our baseline with a ``flat'' (non-hierarchical) decoder, we use a single RNN and split its output to produce per-instrument softmaxes.

\section{Related Work}
\label{sec:related}

A closely related model is the aforementioned recurrent VAE of \cite{bowman2016generating}.
Like ours, their model is effectively a VAE that uses RNNs for both the encoder and decoder.
With careful optimization, \cite{bowman2016generating} demonstrate the ability to generate and interpolate between sentences which have been modeled at the character level.
A very similar model was also proposed by \cite{fabius2014variational}, which was applied with limited success to music.
This approach was also extended to utilize a convolutional encoder and decoder with dilated convolutions in \cite{yang2017improved}.
The primary difference between these models and ours is the decoder architecture; namely, we use a hierarchical RNN.
The flat RNN decoder we use as a baseline in \cref{sec:experiments} exhibits significantly degraded performance when dealing with very long sequences.

Various additional VAE models with autoregressive decoders have also been proposed.
\cite{semeniuta2017hybrid} consider extensions of the recurrent VAE where the RNNs are replaced with feed-forward and convolutional networks.
The PixelVAE \cite{gulrajani2016pixelvae} marries a VAE with a PixelCNN \cite{van2016conditional} and applies the result to the task of natural image modeling. Similarly, the Variational Lossy Autoencoder \cite{chen2016variational} combines a VAE with a PixelCNN/PixelRNN decoder. The authors also consider limiting the power of the decoder and using a more expressive Inverse Autoregressive Flow \cite{kingma2016improved} prior on the latent codes.
Another example of a VAE with a recurrent encoder and decoder is SketchRNN \cite{ha2017neural}, which successfully models sequences of continuously-valued pen coordinates.

The hierarchical paragraph autoencoder proposed in \cite{li2015hierarchical} has several parallels to our work.
They also consider an autoencoder with hierarchical RNNs for the encoder and decoder, where each level in the hierarchy corresponds to natural subsequences in text (e.g., sentences and words).
However, they do not impose any constraints on the latent code, and as a result are unable to sample or interpolate between sequences.
Our model otherwise differs in its use of a flat bidirectional encoder and lack of autoregressive connections in the first level of the hierarchy.

More broadly, our model can be considered in the sequence-to-sequence framework \cite{sutskever2014sequence}, where an encoder produces a compressed representation of an input sequence which is then used to condition a decoder to generate an output sequence. 
For example, the NSynth model learns embeddings by compressing audio waveforms with a downsampling convolutional encoder and then reconstructing audio with a WaveNet-style decoder \cite{engel2017neural}.
Recurrent sequence-to-sequence models are most often applied to sequence \textit{transduction} tasks where the input and output sequences are different.
Nevertheless, sequence-to-sequence autoencoders have been occasionally considered, e.g., as an auxiliary unsupervised training method for semi-supervised learning \cite{dai2015semi} or in the paragraph autoencoder described above.
Again, our approach differs in that we impose structure on the compressed representation (our latent vector) so that we can perform sampling and interpolation.

Finally, there have been many recurrent models proposed where the recurrent states are themselves stochastic latent variables with dependencies across time \cite{chung2015recurrent,bayer2014learning,fraccaro2016sequential}.
A particularly similar example to our model is that of \cite{serban2017hierarchical}, which also utilizes a hierarchical encoder and decoder.
Their model uses two levels of hierarchy and generates a stochastic latent variable for each subsequence of the input sequence.
The crucial difference between this class of models and ours is that we use a \textit{single} latent variable to represent the entire sequence, which allows for creative applications such as interpolation and attribute manipulation.

\section{Experiments}
\label{sec:experiments}

To demonstrate the power of the MusicVAE, we carried out a series of quantitative and qualitative studies on music data.
First, we demonstrate that a simple recurrent VAE like the one described in \cite{bowman2016generating} can effectively generate and interpolate between short sequences of musical notes.
Then, we move to significantly longer note sequences, where our novel hierarchical decoder is necessary in order to effectively model the data.
To verify this assertion, we provide quantitative evidence that it is able to reconstruct, interpolate between, and model attributes from data significantly better than the baseline.
We conclude with a series of listening studies which demonstrate that our proposed model also produces a dramatic improvement in the perceived quality of samples.

\subsection{Data and Training}

For our data source, we use MIDI files, which are a widely-used digital score format.
MIDI files contain instructions for the notes to be played on each individual instrument in a song, as well as meter (timing) information.
We collected \mbox{${\approx}1.5$} million unique files from the web, which provided ample data for training our models.
We extracted the following types of training data from these MIDI files: 2- and 16-bar melodies (monophonic note sequences), 2- and 16-bar drum patterns (events corresponding to playing different drums), and 16-bar ``trio'' sequences consisting of separate streams of a melodic line, a bass line, and a drum pattern.
For further details on our dataset creation process, refer to \cref{sec:data_appendix}.
For ease of comparison, we also evaluated reconstruction quality (as in \cref{sec:reconstruction} below) on the publicly available Lakh MIDI Dataset \cite{raffel2016learning} in \cref{sec:lmd}.

We modeled the monophonic melodies and basslines as sequences of 16th note events.
This resulted in a 130-dimensional output space (categorical distribution over tokens) with 128 ``note-on'' tokens for the 128 MIDI pitches, plus single tokens for ``note-off'' and ``rest''. For drum patterns, we mapped the 61 drum classes defined by the General MIDI standard \cite{international1991general} to 9 canonical classes and represented all possible combinations of hits with 512 categorical tokens.
For timing, in all cases we quantized notes to 16th note intervals, such that each bar consisted of 16 events.
As a result, our two-bar data (used as a proof-of-concept with a flat decoder) resulted in sequences with $T = 32$ and 16-bar data had $T = 256$.
For our hierarchical models, we use $U=16$, meaning each subsequence corresponded to a single bar.

All models were trained using Adam \cite{kingma2014adam} with a learning rate annealed from $10^{-3}$ to $10^{-5}$ with exponential decay rate 0.9999 and a batch size of 512. The 2- and 16-bar models were run for 50k and 100k gradient updates, respectively.
We used a cross-entropy loss against the ground-truth output with scheduled sampling \cite{bengio2015scheduled} for 2-bar models and teacher forcing for 16-bar models.

\subsection{Short Sequences}

As a proof that modeling musical sequences with a recurrent VAE is possible, we first tried modeling 2-bar ($T=32$) monophonic music sequences (melodies and drum patterns) with a flat decoder.
The model was given a tolerance of 48 free bits (${\approx}33.3$ nats) and had the KL cost weight, $\beta$, annealed from 0.0 to 0.2 with exponential rate 0.99999. Scheduled sampling was introduced with an inverse sigmoid rate of 2000.

We found the model to be highly accurate at reconstructing its input (\cref{table:recon} discussed below in \cref{sec:reconstruction}). It was also able to produce compelling interpolations (\cref{fig:mel_2bar_slerp}, \cref{sec:additional_figures}) and samples.
In other words, it learned to effectively use its latent code without suffering from posterior collapse or exposure bias, as particularly evidenced by the relatively small difference in teacher-forced and sampled reconstruction accuracy (a few percent).

Despite this success, the model was unable to reliably reconstruct 16-bar ($T=256$) sequences.
For example, the discrepancy between teacher-forced and sampled reconstruction accuracy increased by more than 27\%.
This motivated our design of the hierarchical decoder described in \cref{sec:decoder}.
In the following sections, we provide an extensive comparison of our proposed model to the flat baseline.

\subsection{Reconstruction Quality}
\label{sec:reconstruction}

\begin{table}[ht]
\vskip -0.03in
\centering
\footnotesize
  \begin{tabular}{lcccc}
    \toprule
    &\multicolumn{2}{c}{Teacher-Forcing} &\multicolumn{2}{c}{Sampling}\\
    \cmidrule(lr){2-3} \cmidrule(lr){4-5}
    Model    & Flat & Hierarchical & Flat & Hierarchical \\
    \midrule
    2-bar Drum          & 0.979 & -     & 0.917 & -     \\
    2-bar Melody        & 0.986 & -     & 0.951 & -     \\
    16-bar Melody       & 0.883 & \textbf{0.919} & 0.620 & \textbf{0.812} \\
    16-bar Drum         & 0.884 & \textbf{0.928} & 0.549 & \textbf{0.879} \\
    Trio (Melody)       & 0.796 & \textbf{0.848} & 0.579 & \textbf{0.753} \\
    Trio (Bass)         & 0.829 & \textbf{0.880} & 0.565 & \textbf{0.773} \\
    Trio (Drums)        & 0.903 & \textbf{0.912} & 0.641 & \textbf{0.863} \\
    \bottomrule
  \end{tabular}
  \caption{Reconstruction accuracies calculated both with teacher-forcing (i.e., next-step prediction) and full sampling. All values are reported on a held-out test set. A softmax temperature of 1.0 was used in all cases, meaning we sampled directly from the logits.}
  \label{table:recon}
\end{table}

To begin, we evaluate whether the hierarchical decoder produces better reconstruction accuracy on 16-bar melodies and drum patterns.
For 16-bar models, we give a tolerance of 256 free bits (${\approx}177.4$ nats) and use $\beta=0.2$.
Table~\ref{table:recon} shows the per-step accuracies for reconstructing the sequences in our test set. 
As mentioned above, we see signs of posterior collapse with the flat baseline, leading to reductions in accuracy of ${\approx}27-32\%$ when teacher forcing is removed for inference. Our hierarchical decoder both increases the next-step prediction accuracy and further reduces the exposure bias by better learning to use its latent code. With the hierarchical model, the decrease in sampling accuracy versus teacher forcing only ranges between ${\approx}5-11\%$.  In general, we also find that the reconstruction errors made by our models are reasonable, e.g., notes shortened by a beat or the addition of notes in the appropriate key.

We also explored the performance of our models on our multi-stream ``trio'' dataset, consisting of 16-bar sequences of melody, bass, and drums.
As with single-stream data, the hierarchical model was able to achieve higher accuracy than the flat baseline while exhibiting a much smaller gap between teacher-forced and sampled performance.

\begin{figure}[htb]
\vskip 0.0in
\begin{center}
\centerline{\includegraphics[width=.8\columnwidth]{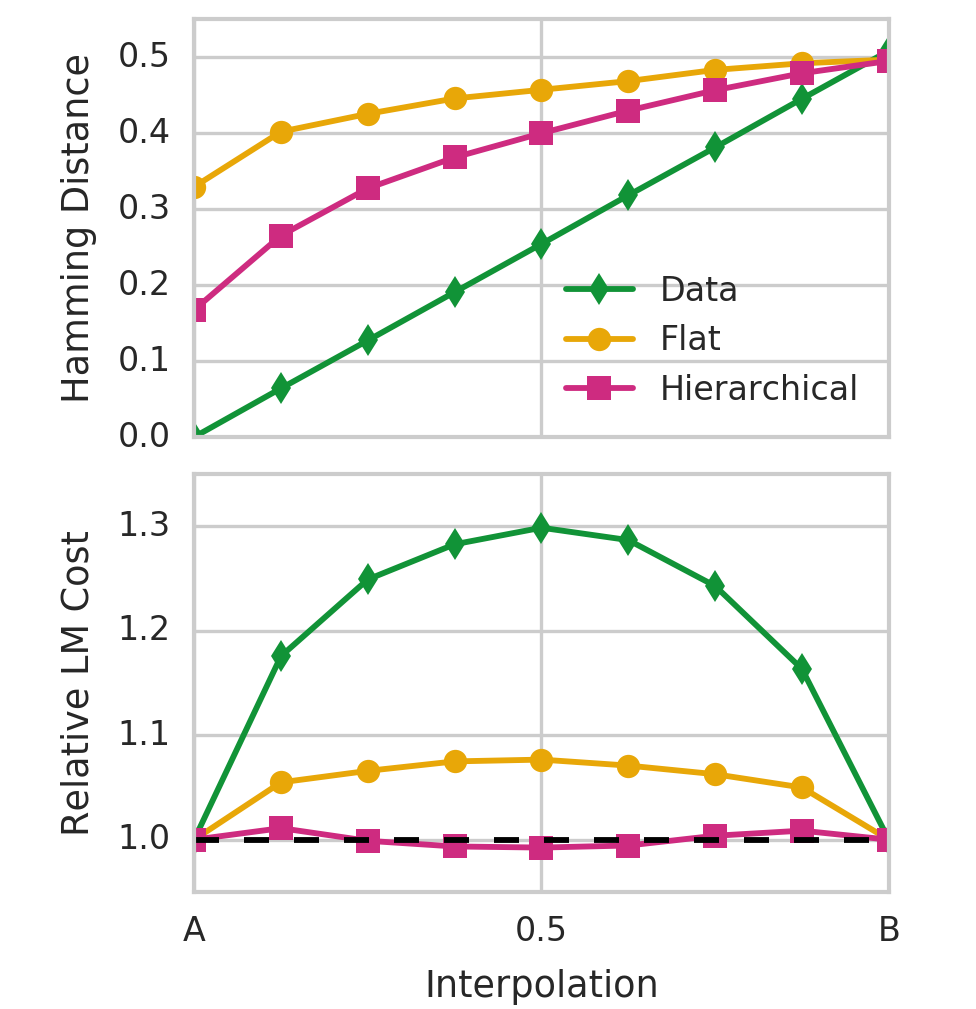}}
\caption{
Latent-space interpolation results. All values are averaged over 1024 interpolated sequences. X-axis denotes interpolation between sequence A to B from left to right. \textit{Top:} Sequence-normalized Hamming distance between sequence A and interpolated points. The distance from B is symmetric to A (decreasing as A increases) and is not shown. \textit{Bottom:} Relative log probability according to an independently-trained 5-gram language model.
}
\label{fig:interpolation}
\end{center}
\vskip -0.4in
\end{figure}

\subsection{Interpolations}

For creative purposes, we desire interpolations that are smoothly varying and semantically meaningful. In \cref{fig:interpolation}, we compare latent-space interpolations from a flat decoder (yellow circles) and hierarchical decoder (magenta squares) to a baseline of naive blending of the two sequences (green diamonds). 
We averaged the behavior of interpolating between 1024 16-bar melodies from the evaluation dataset (A) and 1024 other unique evaluation melodies (B), using a softmax temperature of 0.5 to sample the intermediate sequences.
We constructed baseline ``Data'' interpolations by sampling a Bernoulli random variable with parameter $\alpha$ to choose an element from either sequence $\mathbf{a}$ or $\mathbf{b}$ for each time step, i.e.,  $p(x_t = b_t) = \alpha$, $p(x_t = a_t) = 1- \alpha$.

The top graph of \cref{fig:interpolation} shows that the (sequence length-normalized) Hamming distance, i.e., the proportion of timestep predictions that differ between the interpolation and sequence A, increases monotonically for all methods. The data interpolation varies linearly as expected, following the mean of the Bernoulli distribution. 
The Hamming distances also vary monotonically for latent space interpolations, showing that the decoded sequences morph smoothly to be less like sequence A and more like sequence B. For example, reconstructions don't remain on one mode for several steps and then jump suddenly to another. Samples have a non-zero Hamming distance at the endpoints because of imperfect reconstructions, and the hierarchical decoder has a lower intercept due to its higher reconstruction accuracy. 

For the bottom graph of \cref{fig:interpolation}, we first trained a 5-gram language model on the melody dataset \cite{heafield2011kenlm}.
We show the normalized cost for each interpolated sequence given by $C_\alpha\big/(\alpha C_B + (1-\alpha) C_A)$,
where $C_\alpha$ is the language model cost of an interpolated sequence with interpolation amount $\alpha$, and $C_A$ and $C_B$ are the costs for the endpoint sequences $A$ and $B$.
The large hump for the data interpolation shows that interpolated sequences in data space are deemed by the language model to be much less probable than the original melodies.
The flat model does better, but produces less coherent interpolated sequences than the hierarchical model, which produces interpolations of equal probability to the originals across the entire range of interpolation. 

\cref{fig:interp_mean} shows two example melodies and their corresponding midpoint in MusicVAE space.
The interpolation synthesizes semantic elements of the two melodies: a similar repeating pattern to A, in a higher register with intermittent sparsity like B, and in a new shared musical key. 
On the other hand, the baseline data interpolation just mixes the two resulting in harmonic and rhythmic dissonance (\cref{fig:interpolations_data}, \cref{sec:additional_figures}).


\begin{figure}[t]
\vskip 0.0in
\begin{center}
\centerline{\includegraphics[width=\columnwidth]{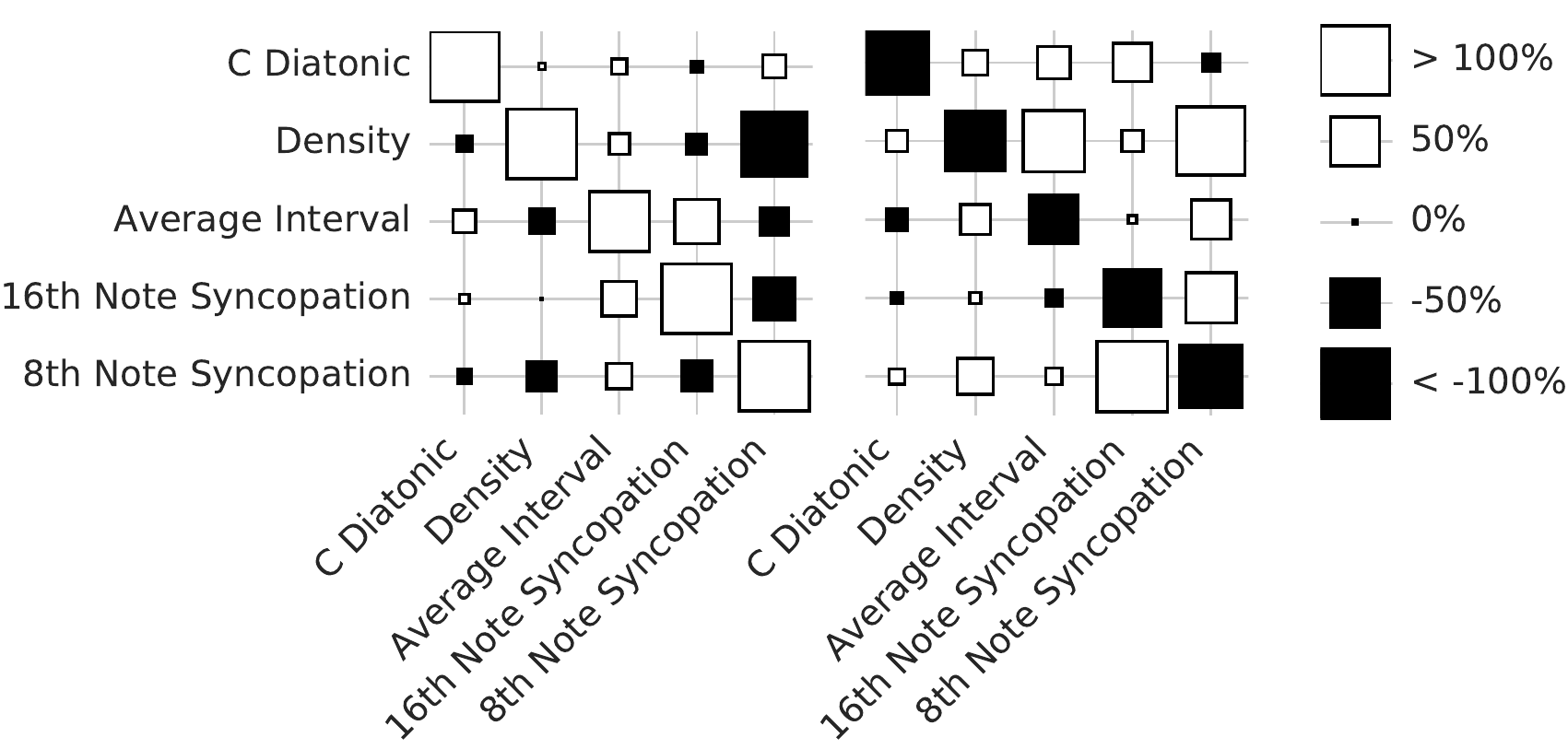}}
\caption{
Adding (left) and subtracting (right) different attribute vectors in latent space and decoding the result produces the intended changes with few side effects.  The vertical axis denotes the attribute vector  applied and the horizontal axis denotes the attribute  measured. See \cref{fig:attribute_long_0,fig:attr_cdiatonic,fig:attr_avgint,fig:attr_16thsync,fig:attr_8thsync} (\cref{sec:additional_figures}) for example piano rolls and \cref{sec:attribute_definitions} for descriptions of each attribute.
}
\label{fig:attr_matrix}
\end{center}
\vskip -0.3in
\end{figure}

\subsection{Attribute Vector Arithmetic}

We can also exploit the structure of the latent space to use ``attribute vectors'' to alter the attributes of a given sequence.
Apart from the score itself, MIDI contains limited annotations \cite{raffel2016extracting}, so we defined five attributes which can be trivially computed directly from the note sequence: C diatonic membership, note density, average interval, and 16th and 8th note syncopation.
See \cref{sec:attribute_definitions} for their full definitions.
Ideally, computing the difference between the average latent vector for sequences which exhibit the two extremes of each attribute and then adding or subtracting it from latent codes of existing sequences will produce the intended semantic manipulation.

To test this, we first measured these attributes in a set of 370k random training examples. For each attribute, we ordered the set by the amount of attribute exhibited, partitioned it into quartiles, and computed an attribute vector by subtracting the mean latent vector of the bottom quartile from the mean latent vector of the top quartile. We then sampled 256 random vectors from the prior, added and subtracted vectors for each attribute, and measured the average percentage change for all attributes against the sequence decoded from the unaltered latent code.
The results are shown in \cref{fig:attr_matrix}.

In general, we find that applying a given attribute vector consistently produces the intended change to the targeted attribute.
We also find cases where increasing one attribute decreases another (e.g.\ increasing density decreases 8th note syncopation), which we believe is largely because our heuristics capture overlapping characteristics.
We are interested in evaluating attribute vector manipulations for labels that are non-trivial to compute (e.g., ornamentation, call/response, etc.) in future work.


\subsection{Listening Tests}

Capturing whether samples from our model sound realistic is difficult to do with purely quantitative metrics. To compare the perceived sample quality of the different models, we therefore carried out listening  studies for melodies, trio compositions, and drum patterns. 

Participants were presented with two 16-bar (${\approx}30$s) compositions that were either sampled from one of the models or extracted from our evaluation dataset. They were then asked which they thought was more musical on a Likert scale. For each study, 192 ratings were collected, with each source involved in 128 pair-wise comparisons. All samples were generated using a softmax temperature of 0.5.

\begin{figure}[t]
\vskip 0.0in
\begin{center}
\centerline{\includegraphics[width=.8\columnwidth]{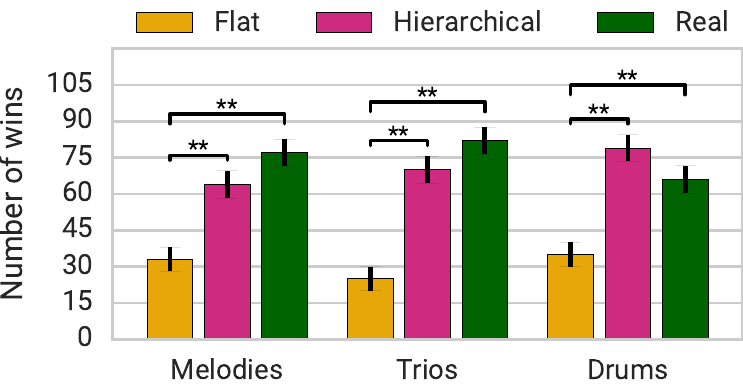}}
\caption{
Results of our listening tests.  Black error bars indicate estimated standard deviation of means. Double asterisks for a pair indicate a statistically significant difference in ranking.
}
\label{fig:listening_tests}
\end{center}
\vskip -0.4in
\end{figure}

\cref{fig:listening_tests} shows the number of comparisons in which a composition from each model was selected as more musical. Our listening test clearly demonstrates the improvement in sample quality gained by using a hierarchical decoder--in all cases the hierarchical model was preferred dramatically more often than the flat model and at the same rate as the evaluation data. In fact, the hierarchical drum model was preferred more often than real data, but the difference is not statistically significant. This was likely due to a listener bias towards variety, as the true drum data, while more realistic, was also more repetitive and perhaps less engaging. 

Further, a Kruskal-Wallis H test of the ratings showed that there was a statistically significant difference between the models: $\chi^2(2) = 37.85, p < 0.001$ for melodies, $\chi^2(2) = 76.62, p < 0.001$ for trios, and $\chi^2(2) = 44.54, p < 0.001$ for drums. A post-hoc analysis using the Wilcoxon signed-rank test with Bonferroni correction showed that participants rated samples from the 3 hierarchical models as more musical than samples from their corresponding flat models with $p < 0.01/3$. Participants also ranked real data as more musical than samples from the flat models with $p < 0.01/3$. There was no significant difference between samples from the hierarchical models and real data.

Audio of some of the examples used in the listening tests is available in the online supplement.\footnoteref{fn:mainonlinesupp}


\section{Conclusion}
\label{sec:conclusion}

We proposed MusicVAE, a recurrent Variational Autencoder which utilizes a hierarchical decoder for improved modeling of sequences with long-term structure.
In experiments on music data, we thoroughly demonstrated through quantitative and qualitative experiments that our model achieves substantially better performance than a flat baseline.
In future work, we are interested in testing our model on other types of sequential data.
To facilitate future research on recurrent latent variable models, we make our code and pre-trained models publicly available.\footnoteref{fn:code}

\section*{Acknowledgements}
\label{sec:acknowledgements}

The authors wish to thank David Ha for inspiration and guidance. We thank Claire Kayacik and Cheng-Zhi Anna Huang for their assistance with the user study analysis. We thank Erich Elsen for additional editing.

\bibliography{example_paper}
\bibliographystyle{icml2018}

\clearpage

\appendix

\section{Dataset Creation Details}
\label{sec:data_appendix}

The datasets were built by first searching the web for publicly-available MIDI files, resulting in ${\approx}1.5$ million unique files. We removed those that were identified as having a non-4/4 time signature and used the encoded tempo to determine bar boundaries, quantizing to 16 notes per bar (16th notes).

For the 2-bar (16-bar) drum patterns, we used a 2-bar (16-bar) sliding window (with a stride of 1 bar) to extract all unique drum sequences (channel 10) with at most a single bar of consecutive rests, resulting in 3.8 million (11.4 million) examples.

For 2-bar (16-bar) melodies, we used a 2-bar (16-bar) sliding window (with a stride of 1 bar) to extract all unique monophonic sequences with at most a single bar of consecutive rests, resulting in 28.0 million (19.5 million) unique examples.

For the trio data, we used a 16-bar sliding window (with a stride of 1 bar) to extract all unique sequences containing an instrument with a program number in the piano, chromatic percussion, organ, or guitar interval, [0, 31], one in the bass interval, [32, 39], and one that is a drum (channel 10), with at most a single bar of consecutive rests in any instrument. If there were multiple instruments in any of the three categories, we took the cross product to consider all possible combinations. This resulted in 9.4 million examples.

In all cases, we reserved a held-out evaluation set of examples which we use to report reconstruction accuracy, interpolation results, etc.

\section{Lakh MIDI Dataset Results}
\label{sec:lmd}

For easier comparison, we also trained our 16-bar models on the publicly available Lakh MIDI Dataset (LMD) \cite{raffel2016learning}, which makes up a subset of the our dataset described above. We extracted 3.7 million melodies, 4.6 million drum patterns, and 116 thousand trios from the full LMD. The models were trained with the same hyperparameters as were used for the full dataset. 

We first evaluated the LMD-trained melody model on a subset of the full evaluation set made by excluding any examples in the LMD train set. We found less than a 1\% difference in reconstruction accuracies between the LMD-trained and original model.

In \cref{table:lmd} we report the reconstruction accuracies for all 3 16-bar models trained and evaluated on LMD. While the accuracies are slightly higher than \cref{table:recon}, the same conclusions regarding the relative performance of the models hold.

\begin{table}[ht]
\centering
\footnotesize
  \begin{tabular}{lcccc}
    \toprule
    &\multicolumn{2}{c}{Teacher-Forcing} &\multicolumn{2}{c}{Sampling}\\
    \cmidrule(lr){2-3} \cmidrule(lr){4-5}
    Model    & Flat & Hierarchical & Flat & Hierarchical \\
    \midrule
    16-bar Melody       & 0.952 & \textbf{0.956} & 0.685 & \textbf{0.867} \\
    16-bar Drum         & 0.937 & \textbf{0.955} & 0.794 & \textbf{0.908} \\
    Trio (Melody)       & 0.866 & \textbf{0.868} & 0.660 & \textbf{0.760} \\
    Trio (Bass)         & 0.906 & \textbf{0.912} & 0.651 & \textbf{0.782} \\
    Trio (Drums)        & 0.943 & \textbf{0.946} & 0.641 & \textbf{0.895} \\
    \bottomrule
  \end{tabular}
  \caption{Reconstruction accuracies for the Lakh MIDI Dataset calculated both with teacher-forcing (i.e., next-step prediction) and full sampling. All values are reported on a held-out test set. A softmax temperature of 1.0 was used in all cases, meaning we sampled directly from the logits.}
  \label{table:lmd}
\end{table}

\section{Attribute Definitions}
\label{sec:attribute_definitions}

The following definitions were used to measure the amount of each attribute.

\subsection*{C Diatonic}
The fraction of notes in the note sequence whose pitches lay in the diatonic scale on C (A-B-C-D-E-F-G, i.e., the ``white keys").

\subsection*{Note Density}
The number of note onsets in the sequence divided by the total length of the sequence measured in 16th note steps.

\subsection*{Average Interval}
The mean absolute pitch interval between consecutive notes in a sequence.

\subsection*{16th Note Syncopation}
The fraction of (16th note) quantized note onsets landing on an odd 16th note position (1-indexed) with no note onset at the previous 16th note position. 

\subsection*{8th Note Syncopation}
The fraction of (16th note) quantized note onsets landing on an odd 8th note position (1-indexed) with no note onset at either the previous 16th or 8th note positions.

\section{Audio Samples}
Synthesized audio for all examples here and in the main text can be found in the online supplement.\footnote{\label{fn:onlinesupp}\url{https://goo.gl/magenta/musicvae-examples}}

\section{Additional Figures and Samples}
\label{sec:additional_figures}

Subsequent pages include additional figures, referenced from the main text.

\clearpage

\begin{figure}[htb]
\vskip 0.2in
\begin{center}
\centerline{\includegraphics[width=\columnwidth]{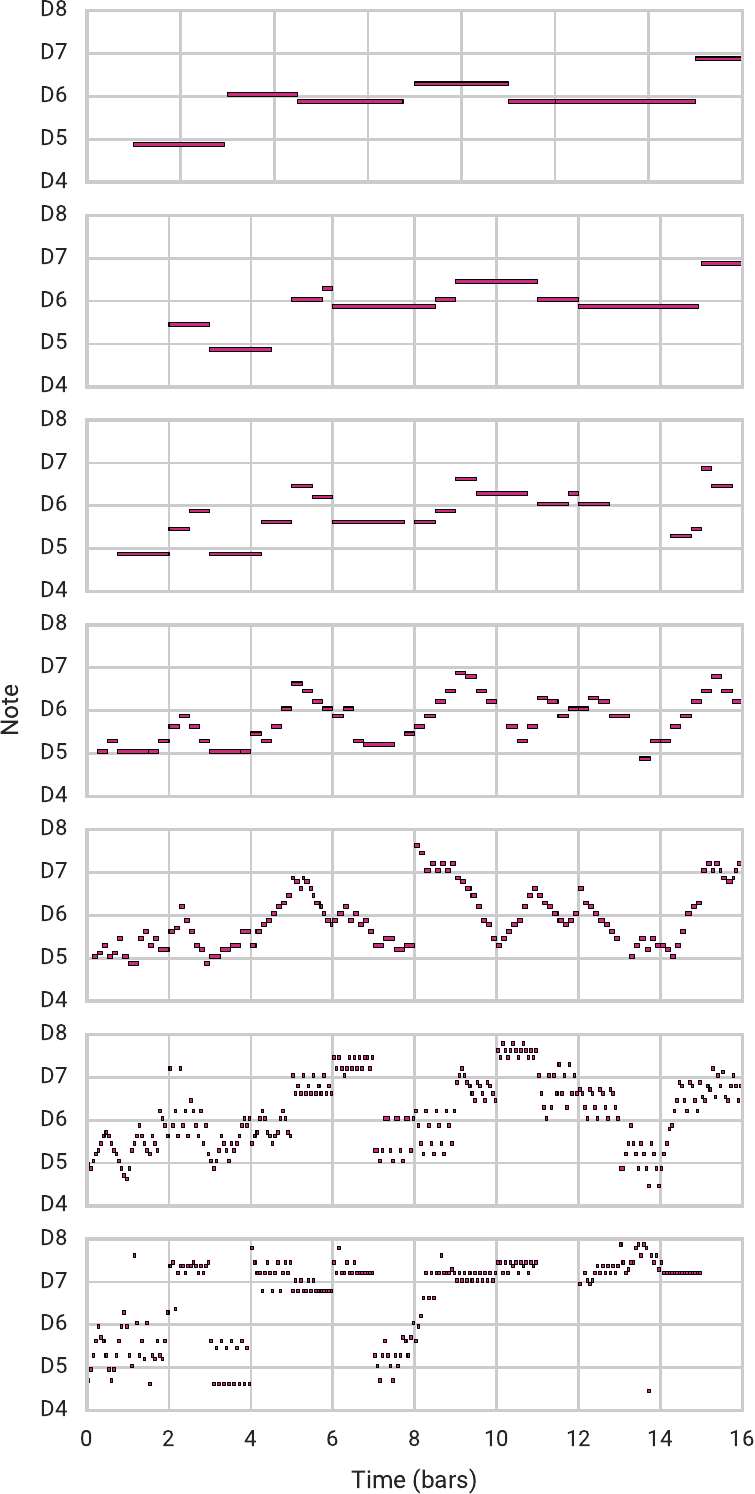}}
\caption{
Varying the amount of the ``Note Density'' attribute vector. The amount varies from -1.5 to 1.5 in steps of 0.5, with the central sequence corresponding to no attribute vector.
Audio for this example is available in the online supplement.\footnoteref{fn:onlinesupp}
}
\label{fig:attribute_long_0}
\end{center}
\vskip -0.2in
\end{figure}

\begin{figure}[htb]
\vskip 0.2in
\begin{center}
\centerline{\includegraphics[width=\columnwidth]{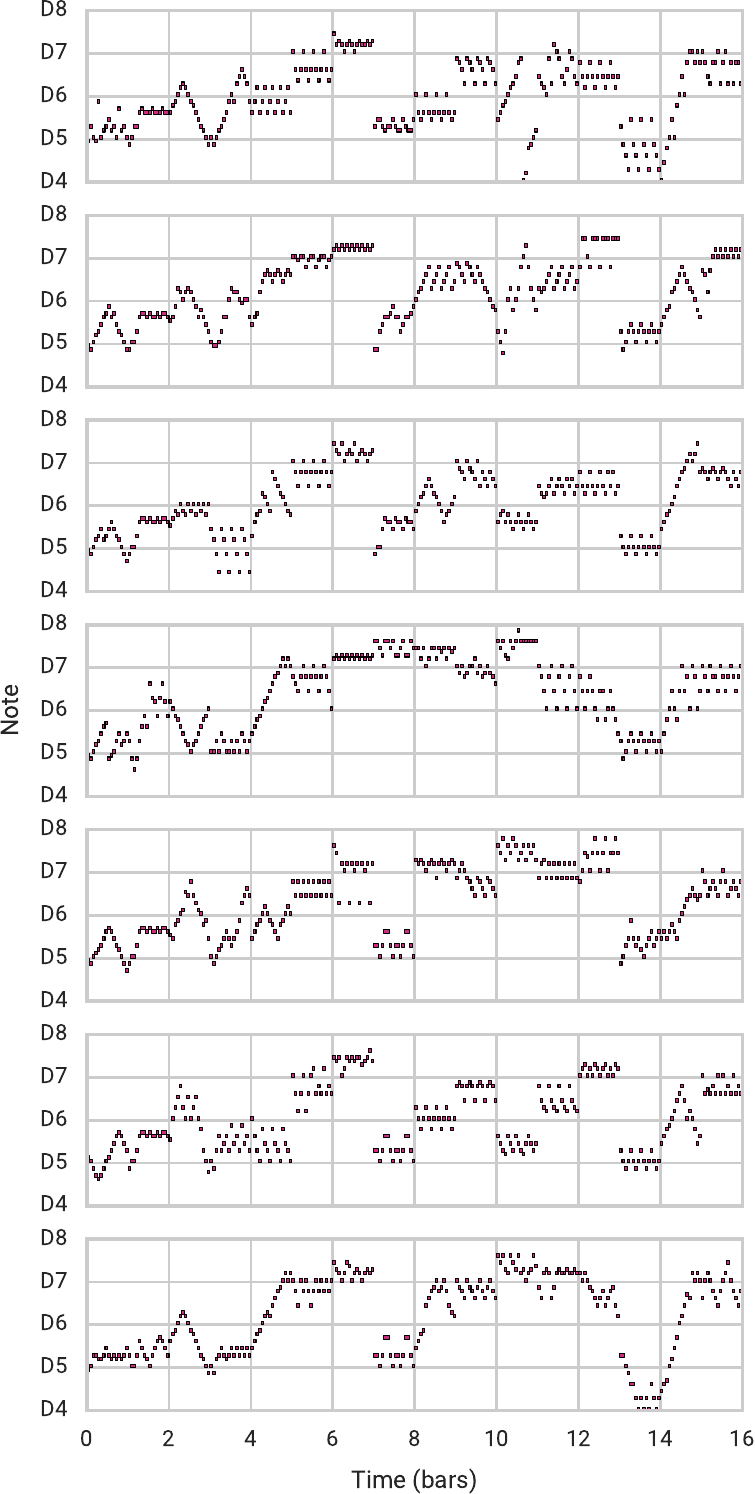}}
\caption{
Additional resamplings of the same latent code (corresponding to the second-to-the-bottom in \cref{fig:attribute_long_0}). While semantically similar, the specific notes vary due to the sampling in the autoregressive decoder.
Audio for this example is available in the online supplement.\footnoteref{fn:onlinesupp}
}
\label{fig:attribute_resample_0}
\end{center}
\vskip -0.2in
\end{figure}

\clearpage

\begin{figure}[t]
\begin{center}
\centerline{\includegraphics[width=\columnwidth]{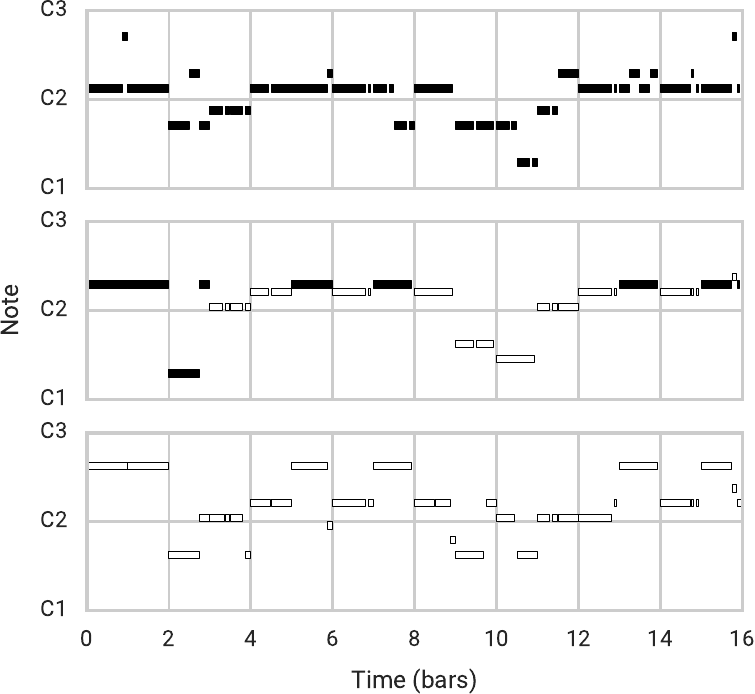}}
\caption{
Subtracting (top) and adding (bottom) the ``C Diatonic'' attribute vector from the note sequence in the middle. For ease of interpretation, notes in the C diatonic scale are shown in white and notes outside the scale are shown in black.
Audio for this example is available in the online supplement.\footnoteref{fn:onlinesupp}
}
\label{fig:attr_cdiatonic}
\end{center}
\end{figure}

\begin{figure}[b]
\begin{center}
\centerline{\includegraphics[width=\columnwidth]{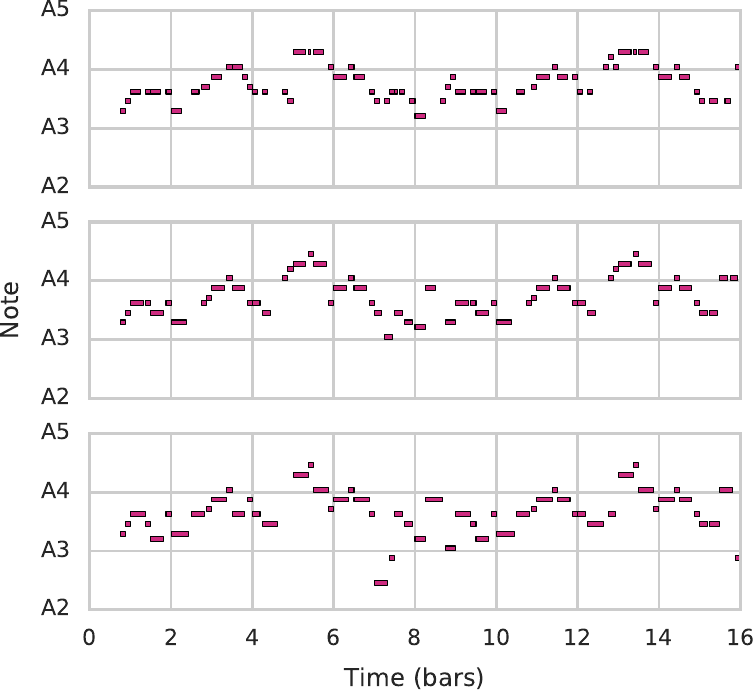}}
\caption{
Subtracting (top) and adding (bottom) the ``Average Interval'' attribute vector from the note sequence shown in the middle.
Audio for this example is available in the online supplement.\footnoteref{fn:onlinesupp}
}
\label{fig:attr_avgint}
\end{center}
\end{figure}

\begin{figure}[t]
\vskip 0.1in
\begin{center}
\centerline{\includegraphics[width=\columnwidth]{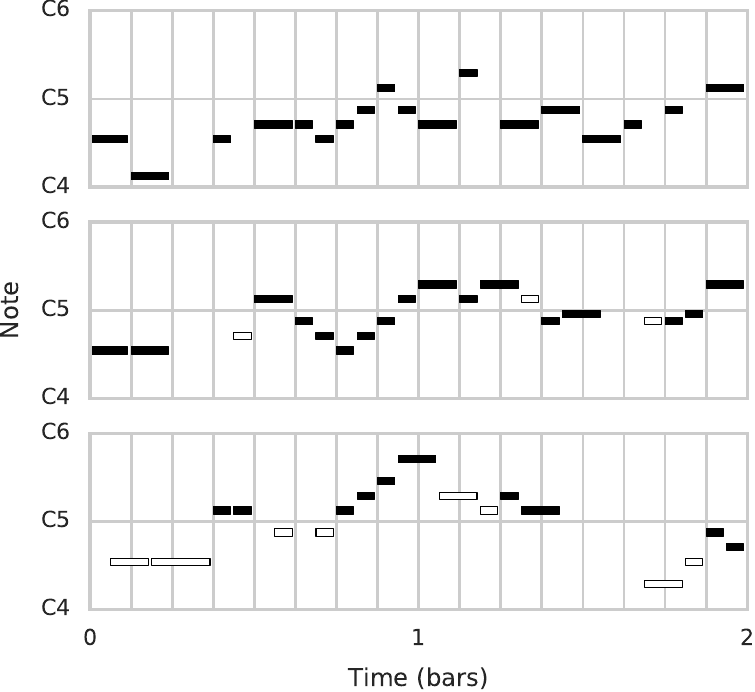}}
\caption{
Subtracting (top) and adding (bottom) the ``16th Note Syncopation'' attribute vector from the note sequence in the middle.
For ease of interpretation, only the first 2 of each sequence's 16 bars are shown.
Vertical lines indicate 8th note boundaries.
White and black indicate syncopated and non-syncopated notes, respectively.
Audio for this example is available in the online supplement.\footnoteref{fn:onlinesupp}
}
\label{fig:attr_16thsync}
\end{center}
\vskip -0.1in
\end{figure}

\begin{figure}[b]
\vskip 0.25in
\begin{center}
\centerline{\includegraphics[width=\columnwidth]{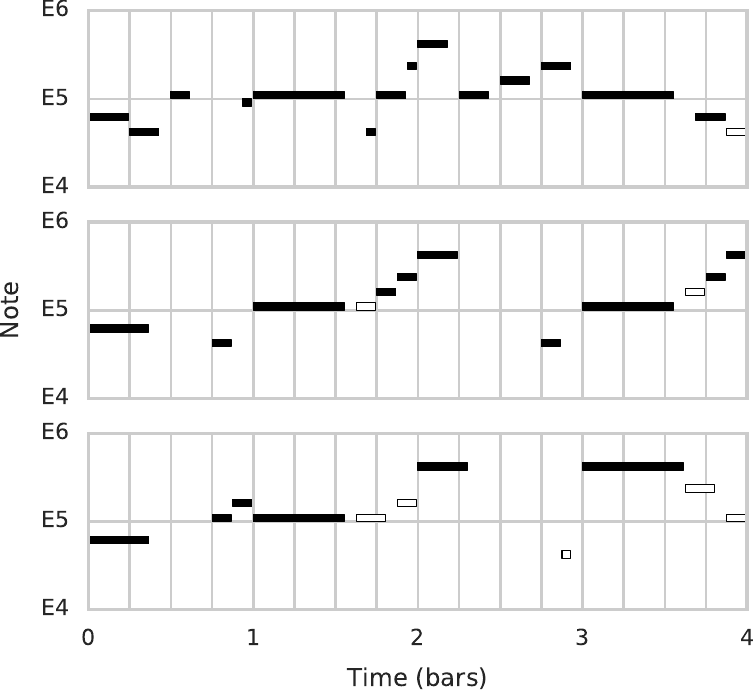}}
\caption{
Subtracting (top) and adding (bottom) the ``8th Note Syncopation'' attribute vector from the note sequence in the middle.
For ease of interpretation, only the first 4 of each sequence's 16 bars are shown.
Vertical lines indicate quarter note boundaries.
White and black indicate syncopated and non-syncopated notes, respectively.
Audio for this example is available in the online supplement.\footnoteref{fn:onlinesupp}
}
\label{fig:attr_8thsync}
\end{center}
\vskip -0.25in
\end{figure}

\begin{figure}[htb]
\vskip 0.2in
\begin{center}
\centerline{\includegraphics[width=\columnwidth]{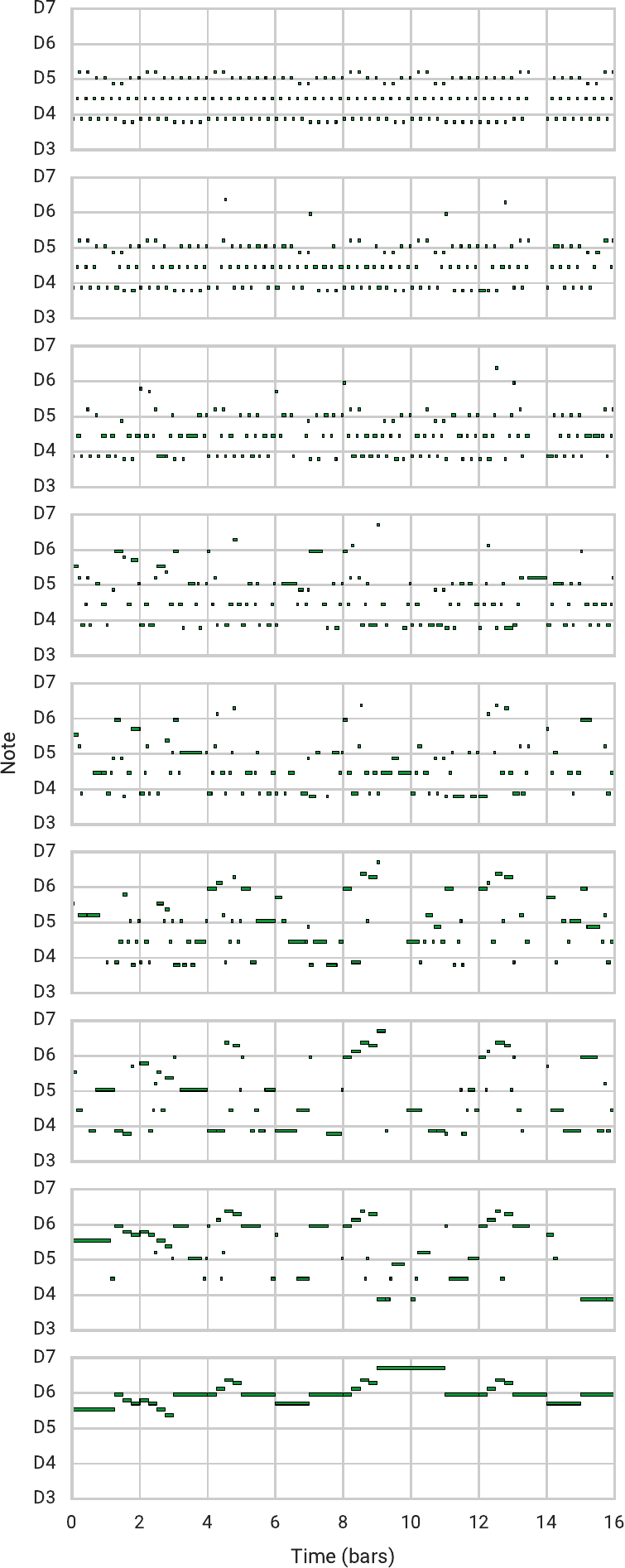}}
\caption{
Interpolating between the top and bottom sequence in data space.
Audio for this example is available in the online supplement.\footnoteref{fn:onlinesupp}
}
\label{fig:interpolations_data}
\end{center}
\vskip -0.2in
\end{figure}

\begin{figure}[htb]
\vskip 0.2in
\begin{center}
\centerline{\includegraphics[width=\columnwidth]{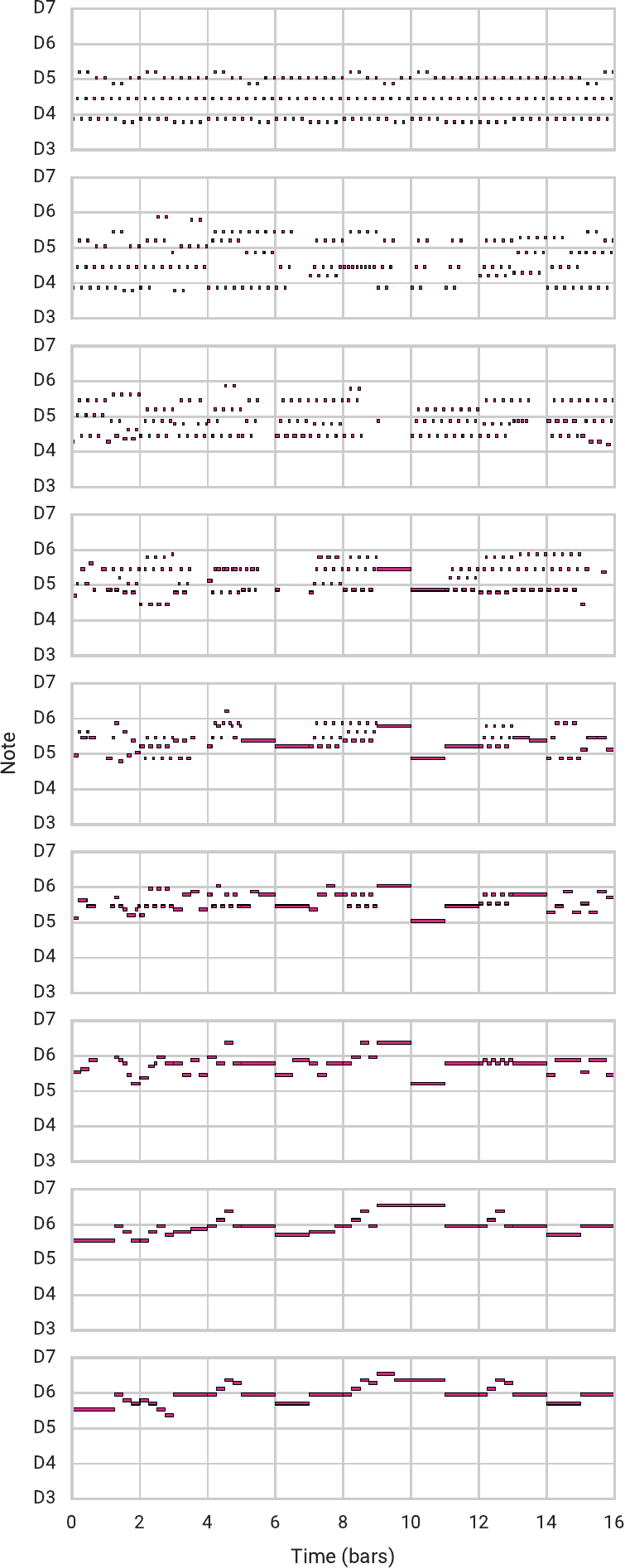}}
\caption{
Interpolating between the top and bottom sequence (same as \cref{fig:interpolations_data}) in MusicVAE's latent space.
Audio for this example is available in the online supplement.\footnoteref{fn:onlinesupp}
}
\label{fig:interpolations_hierarchical}
\end{center}
\vskip -0.2in
\end{figure}

\begin{figure*}[htb]
\vskip 0.2in
\begin{center}
\centerline{\includegraphics[width=\textwidth]{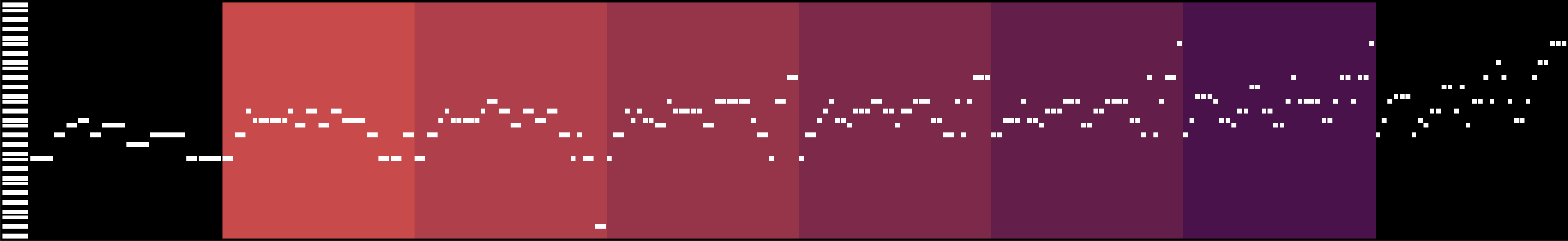}}
\caption{
Example interpolation in the 2-bar melody MusicVAE latent space. Vertical axis is pitch (from $A_3$ to $C_8$) and horizontal axis is time. We sampled 6 interpolated sequences between two test-set sequences on the left and right ends. Each 2-bar sample is shown with a different background color. Audio of an extended, 13-step interpolation between these sequences is available in the online supplement.\footnoteref{fn:onlinesupp}
}
\label{fig:mel_2bar_slerp}
\end{center}
\vskip -0.2in
\end{figure*}

\begin{figure*}[htb]
\vskip 0.2in
\begin{center}
\centerline{\includegraphics[width=\textwidth]{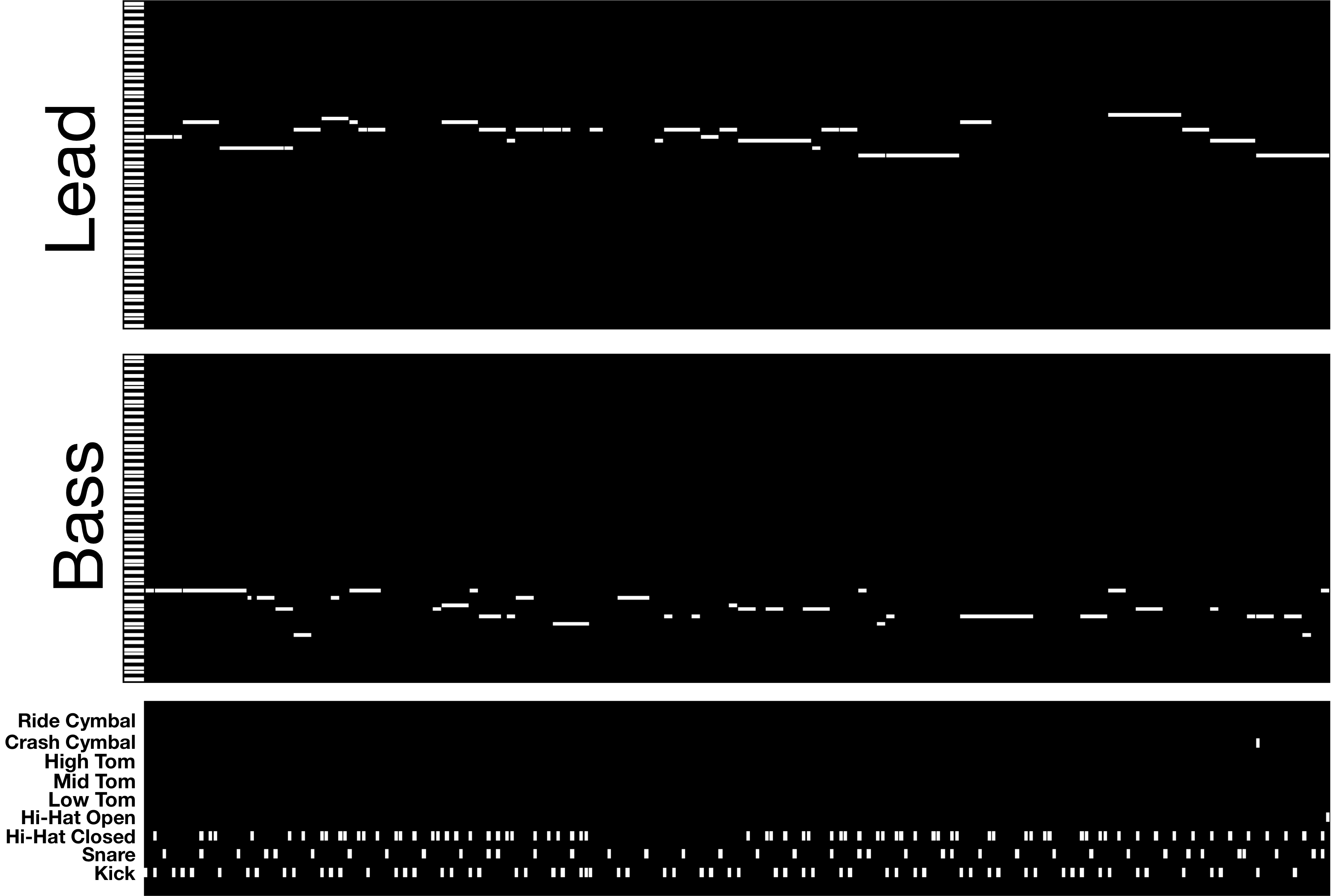}}
\caption{
Selected example 16-bar trio sample generated by MusicVAE.
Audio for this and other samples is available in the online supplement.\footnoteref{fn:onlinesupp}
}.
\label{fig:trio_16bar}
\end{center}
\vskip -0.2in
\end{figure*}

\end{document}